\documentclass[twoside]{article}

\usepackage{algorithm}
\usepackage{algorithmic}
\newcommand{\softmax}{\mbox{softmax}}
\newcommand{\argmax}{\mbox{argmax}}
\usepackage{multirow}
\usepackage{graphicx}
\newcommand{\ie}{i.e.}

\usepackage{amsfonts}
\graphicspath{{./figures}}

\usepackage[accepted]{aistats2024}

\usepackage[round]{natbib}

\begin{document}

\runningtitle{Cross-model Mutual Learning for Exemplar-based Medical Image Segmentation}

\twocolumn[

\aistatstitle{Cross-model Mutual Learning for Exemplar-based \\ Medical Image Segmentation}

\aistatsauthor{ Qing En$^*$ \qquad 
Yuhong Guo$^{*\dagger}$ }

\aistatsaddress{ $^*$School of Computer Science, Carleton University, Ottawa, Canada  \\ 
$^\dagger$Canada CIFAR AI Chair, Amii, Canada \\
qingen@cunet.carleton.ca,
yuhong.guo@carleton.ca
} ]

\begin{abstract}
Medical image segmentation typically demands extensive dense annotations for model training, which is both time-consuming and skill-intensive.
To mitigate this burden, exemplar-based medical image segmentation methods have been introduced to achieve effective training with only one annotated image.
In this paper, we introduce a novel Cross-model Mutual learning framework for Exemplar-based Medical image Segmentation (CMEMS), which leverages two models to mutually excavate implicit information from unlabeled data at multiple granularities.
CMEMS can eliminate confirmation bias and enable collaborative training to learn complementary information by enforcing consistency at different granularities across models. Concretely, cross-model image perturbation based mutual learning
is devised by using weakly perturbed images to generate high-confidence pseudo-labels, supervising predictions of strongly perturbed images across models. 
This approach enables joint pursuit of prediction consistency at the image granularity.
Moreover, cross-model multi-level feature perturbation based mutual learning is designed by letting pseudo-labels supervise predictions from perturbed multi-level features with different resolutions, which can broaden the perturbation space and enhance the robustness of our framework.
CMEMS is jointly trained using exemplar data, synthetic data, and unlabeled data in an end-to-end manner.
Experimental results on two medical image datasets indicate that the proposed CMEMS outperforms the state-of-the-art segmentation methods with extremely limited supervision.
\end{abstract}

\section{INTRODUCTION}
Medical image analysis has gained significant attention in clinical diagnostics and complementary medicine due to the rapid advancements in medical image technology \citep{duncan2000medical,anwar2018medical}.
In particular, medical image segmentation is a fundamental and challenging task that involves determining the category of each pixel in medical images \citep{sharma2010automated,ronneberger2015u}.
While existing fully supervised methods have produced satisfactory results, obtaining numerous fine-grained annotations is both time-consuming and labour-intensive, which is a major barrier to the advancement of medical image analysis.
Several approaches have been proposed to alleviate this burden by using limited annotations to achieve complex tasks and overcome the limitations of fully supervised methods \citep{qian2019weakly,tang2018regularized,lin2016scribblesup}.

Although existing methods have made considerable progress, they still struggle to overcome the barriers presented by labeling. 
They typically require either class and location information of each image or a portion of fine annotations during the training and testing phases \citep{roy2020squeeze,tang2021recurrent,luo2021semi,seibold2021reference,wu2021collaborative}.
Unlike the above methods, exemplar learning-based medical image segmentation can enjoy the ability to complete the segmentation task with only one expert-annotated image, owing to the unique properties of medical images \citep{en2022exemplar}.
Additionally, this technique can utilize new unlabeled data to continuously improve the model's effectiveness.
Consequently, we concentrate on segmenting medical images in the exemplar-learning scenario, in which the only annotated image serves as an exemplar containing all organ categories present in the dataset.

Recently, an exemplar-learning based method called ELSNet \citep{en2022exemplar} has been proposed by synthesizing annotated images and applying a contrastive approach to learn 
segmentation models from exemplars.
However, the use of static pseudo-labels restricts it from taking full advantage of the knowledge of unlabelled data.
Furthermore, it is inappropriate to directly apply existing semi-supervised methods to the ELSNet 
as they ignore the diverse granularity of consistency information and do not leverage 
the complementary knowledge offered by multiple models simultaneously, which can lead to a dilemma in the exemplar learning scenario.

The fundamental challenges of exemplar-learning based medical semantic segmentation can be attributed to the following two aspects:
(1) With only one annotated exemplar image, it is difficult to train the model effectively when inaccurate pseudo-labels produced by unlabeled images hinder the training process.
(2) The lack of guidance from complementary information, combined with insufficient label diversity and unsatisfactory quality of pseudo labels, 
also results in confirmation bias and aggravates the difficulty of the task.
Humans have the ability to exhibit different patterns of behaviour, whether in the face of integral or separate features \citep{zhu2011co}.
Inspired by the idea, we propose integrating cross-model learning at different granularities to obtain consistency and complementary information across various levels of detail.

\begin{figure}
\centering
  \includegraphics[width=0.8\linewidth,height=32mm]{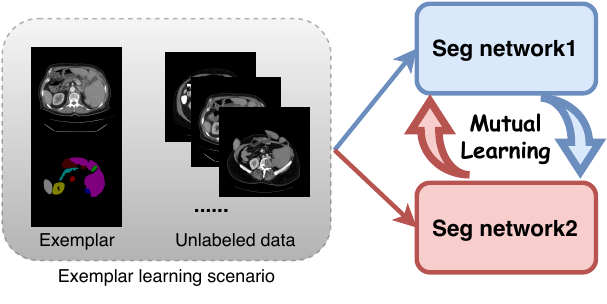}\\
\caption{
Illustration of the proposed idea.
The proposed CMEMS leverages two mutual learning models to excavate implicit information from unlabeled data for exemplar-based medical image segmentation.
}
  \label{fig:idea}
\end{figure}

In this paper, we propose a novel Cross-Model Mutual Learning framework for Exemplar-based Medical image Segmentation (CMEMS), as shown in Figure \ref{fig:idea}.
The core of this framework is to leverage two mutual segmentation models that exploit implicit information from unlabeled data at various granularity levels to produce more precise pseudo-labels, thereby enhancing the performance of both models through mutual supervision.
CMEMS can alleviate confirmation bias and enable the acquisition of complementary information by promoting consistency across various granularities of unlabeled images and facilitating collaborative training of multiple models \citep{arazo2020pseudo}.
Specifically, we devise a cross-model image perturbation based mutual learning mechanism
that leverages high-confidence pseudo-labels obtained from weakly perturbed unlabeled images by one model to supervise the predictions of strongly perturbed unlabeled images generated by the other model.
In this case, the two models can jointly pursue prediction consistency at the image granularity by computing the cross-model image perturbation loss.
Moreover, we present a cross-model multi-level feature perturbation based mutual learning mechanism 
by letting pseudo-labels supervise predictions of perturbed multi-level features to broaden the perturbation space and strengthen the robustness of our framework.
This is achieved by computing the cross-model multi-layer feature perturbation loss, which can enable the framework to maintain feature-level consistency across models.
Finally, the supervised segmentation losses calculated from the exemplar 
and the generated synthetic dataset 
are combined with the two losses mentioned above to optimize the proposed CMEMS collaboratively.
Extensive experimental results demonstrate the effectiveness of the proposed CMEMS framework.
In summary, the key contributions of our paper are as follows:
\begin{itemize}
\item 
We propose a novel CMEMS framework for exemplar-based medical image segmentation by leveraging two mutual learning models to excavate implicit information from unlabeled data at different granularities.

\item 
We devise a cross-model image perturbation based mutual learning mechanism 
and a cross-model multi-level feature perturbation based mutual learning mechanism
by enforcing consistency across unlabeled images and features to alleviate confirmation bias and enable the acquisition of complementary information.

\item 
Experimental results demonstrate that the proposed CMEMS framework achieves state-of-the-art performance on the Synapse and ACDC medical image datasets in exemplar learning scenarios.
\end{itemize}

\begin{figure*}
\centering
  \includegraphics[width=1\linewidth,height=55mm]{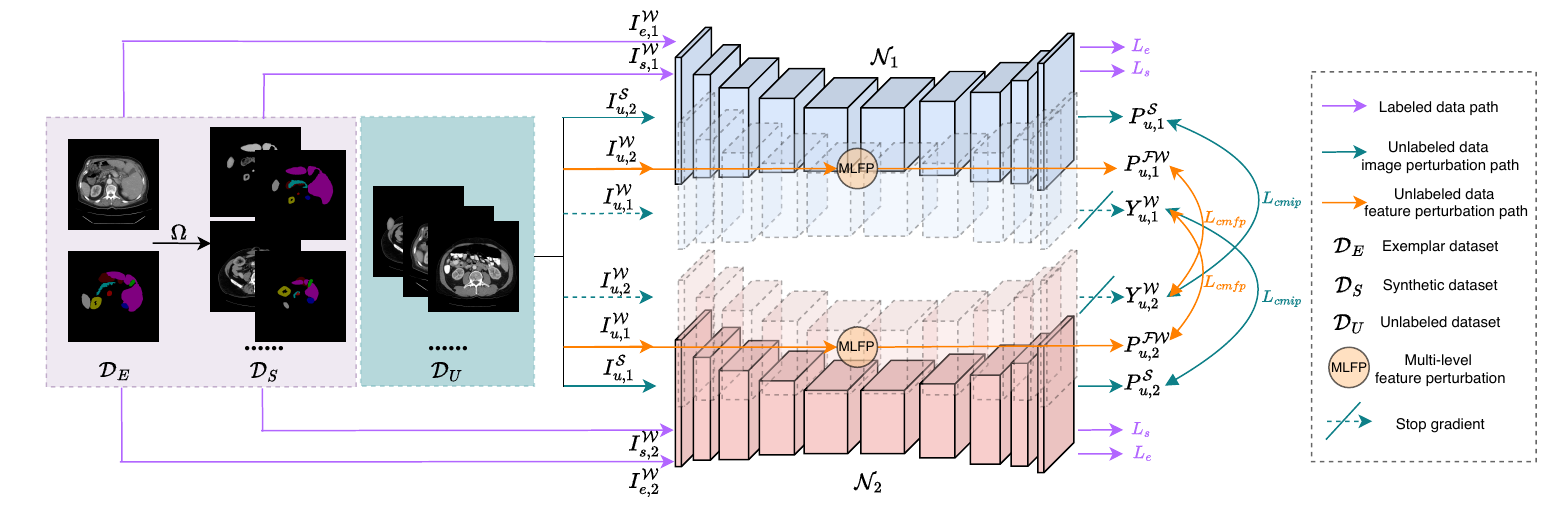}\\
\caption{
An overview of the proposed CMEMS.
Firstly, a synthetic dataset $\mathcal{D}_{S}$ is generated from an exemplar dataset $\mathcal{D}_{E}$ using the $\Omega$ method. 
Both datasets are then fed into segmentation networks $\mathcal{N}_{m} (m \in \{1,2\})$ to calculate $L_{e}$ and $L_{s}$.
Next, the unlabeled image $I_{u}$ is fed into two segmentation networks using weak and strong perturbations, respectively, to calculate $L_{cmip}$ by cross-model image perturbation based mutual learning and to calculate $L_{cmfp}$ via cross-model multi-level feature perturbation based mutual learning.
Finally, the two segmentation networks are optimized collaboratively by computing all loss functions.
}
\label{fig:CMEMS}
\end{figure*}
\section{RELATED WORKS}
\subsection{Medical Image Segmentation}
The medical image segmentation task aims to identify the organ or lesion area of the input medical image.
CNN-based methods \citep{milletari2016v,ronneberger2015u} and transformer-based methods \citep{chen2021transunet,wang2022mixed,li2022tfcns} have been widely used to solve this problem with promising results.
UNet \citep{ronneberger2015u} is the most widely used encoder-decoder structured CNN model that can efficiently solve medical image segmentation tasks.
Its variants have produced better segmentation results \citep{zhang2019net,huang2020unet,rahman2023medical}.
Besides, transformer-based methods have achieved notable success in medical image segmentation by capturing long-range dependencies \citep{cao2022swin,wang2022mixed}.
However, the abovementioned methods depend on many labeled images, whereas we intend to address the challenging 
scenario where only one labeled image is available.

\subsection{Semi-Supervised Semantic Segmentation}
Semi-supervised semantic segmentation has been extensively investigated and gained success \citep{zoupseudoseg,ouali2020semi,chen2021semi,wang2022semi,yuan2023semi,yang2023revisiting}.
Recent attempts have focused on consistency regularization \citep{french2019semi,chen2021semi,hu2021semi,lai2021semi} and entropy minimization \citep{yuan2021simple,yang2022st++,guan2022unbiased}.
Some methods \citep{luo2022semi,chen2021semi,zheng2022uncertainty} use the same images for co-training or uncertainty consistency across models but overlook varying levels of perturbations.
Instead, we aim to solve the problem of segmenting medical images based on exemplar learning in a unique, more challenging experimental setting than natural images.

\subsection{Medical Image Segmentation with Limited Supervision}
Many methods have been proposed to achieve medical image analysis using limited supervision.
Some works utilized the mean-teacher model \citep{yu2019uncertainty} and incorporated the self-ensembling framework \citep{cui2019semi,li2020transformation,seibold2021reference,you2022momentum} to explore the prediction information of the unlabeled images.
Several uncertainty constraints and co-training methods \citep{luo2021efficient,wu2022mutual,luo2022semi,xia2020uncertainty} have been attempted to enable the model to generate invariant results by minimizing the discrepancy of outputs.
One step further, the first and so far only exemplar learning method for medical image segmentation, ELSNet \citep{en2022exemplar}, has been proposed, which uses one annotated image for medical image semantic segmentation.
Unlike ELSNet's single-model approach, our framework uses two mutual learning models to guide the network away from incorrect directions. 
Besides, while ELSNet relies on static pseudo-labels, our framework dynamically generates them through image and multi-level feature perturbation, exploiting implicit information from unlabeled data at various granularity levels for more precise labeling.

\section{METHOD}
\label{sec:method}
In this section, we present the proposed CMEMS framework for exemplar-based medical image segmentation.
We first introduce the experimental setup and architecture of the proposed CMEMS.
Next, we briefly introduce exemplar-based data synthesis.
We then present the cross-model mutual learning framework for exemplar-based medical image segmentation, including cross-model image perturbation
based mutual learning and cross-model multi-level feature perturbation
based mutual learning.
Finally, we present the collaborative optimization procedure.

\subsection{Overview}
\label{sec:overview}
In the exemplar learning scenario, a single exemplar image is defined as an image containing one segmentation instance for each category present in the dataset.
We aim to train two good segmentation models using a single labeled training image (\ie\ exemplar) and $T$ 
unlabeled training images, represented as $\mathcal{D}_{E} = (I_{e}, Y_{e})$ and $\mathcal{D}_{U} = \{(I_{u}^t)\}_{t=1}^{T}$, respectively.
Let $I\in \mathbb{R}^{1\times H\times W}$ 
denote a 2D input image and  $Y_{e}\in \{0, 1\}^{K\times H\times W}$ denote the corresponding 
segmentation labels, 
where $K$ is the number of classes (\ie, categories of organs), and $H$ and $W$ represent the height and width of the input image, respectively.

The architecture of the proposed CMEMS is presented in Figure \ref{fig:CMEMS}, comprising two segmentation networks with identical structures but distinct parameters.
We initially generate a synthetic dataset $\mathcal{D}_{S}$ from the exemplar $\mathcal{D}_{E}$ to enhance the diversity of the training set.
Next, given the unlabeled dataset $\mathcal{D}_{U}$, cross-model image perturbation learning is accomplished by using pseudo-labels produced from weakly perturbed images by one model to supervise predictions of strong perturbed unlabeled images by the other model, which can make two models jointly pursue prediction consistency at the image granularity.
Furthermore, pseudo-labels are used to supervise the predictions obtained from perturbed multi-level features in cross-model multi-level feature perturbation learning.
This broadens the perturbation space and enhances the robustness of the framework at the feature granularity.
Finally, both segmentation networks are jointly optimized using the exemplar, synthetic, and unlabeled datasets.
Each of the segmentation networks, denoted as $\mathcal{N}_{m} (m \in {1,2})$, comprises of an encoder 
$f_m: 
\mathbb{R}^{1\times H\times W} \to \mathbb{R}^{c\times h\times w}$ and a decoder 
$g_m: 
\mathbb{R}^{c\times h\times w} \to \mathbb{R}^{K\times H\times W}$, 
where $c$, $h$, and $w$ denote the number of channels, height, and width of the encoded features, respectively. 
The proposed CMEMS framework is trained end-to-end.

\subsection{Exemplar-based Data Synthesis}
\label{sec:Exemplar-based data synthesis}
We first employ an efficient method to synthesize segmentation instances of each label category from an exemplar into various backgrounds to augment the diversity of annotated data \citep{en2022exemplar}.
For an exemplar image $I_{e}$ and its annotation $Y_{e}$, we use a series of geometric transformations $\mathcal{G}$ and intensity transformations $\mathcal{I}$ to generate a synthetic image $I_{s}$ and its annotation $Y_{s}$ through a copy-and-paste strategy as follows:
\begin{equation}
\begin{split}
(I_{s}, Y_{s}) = \Omega(\mathcal{G}(\mathcal{I}(I_{e})), \mathcal{G}(Y_{e}), \mathcal{G}(\mathcal{I}(I_{b})) ).
\end{split}
\label{transformation}
\end{equation}
where $\Omega$ denotes the operation of copying, transforming and pasting the exemplar organ onto the background image $I_{b}$ and generating the corresponding annotation.
Using this operation, 
we construct a synthetic dataset $\mathcal{D}_{S} = \{(I_{s}^b, Y_{s}^b)\}_{b=1}^{B}$, where $I_{s}^b\in \mathbb{R}^{1\times H\times W}$, $Y_{s}^b\in \{0, 1\}^{K\times H\times W}$, and $B$ is the number of synthetic images.

\subsection{Cross-model Mutual Learning for Exemplar-based Medical Image Segmentation}
\label{sec:CMEMS}
Although the synthetic dataset can increase the diversity of annotated data, it still provides limited supervision 
information. 
It is desirable to utilize the abundant unlabeled images with predicted pseudo-labels to assist the learning of accurate segmentation models.
Despite ELSNet \citep{en2022exemplar} attempts to 
increase the discriminative ability of the segmentation network to obtain better pseudo-labels, 
the model is still negatively impacted by noise from static pseudo-labels 
and its own confirmation bias.
Therefore, we propose to leverage
two mutual learning segmentation networks to dynamically generate pseudo-labels 
through image and feature perturbation learning across models, aiming to learn complementary information and eliminate confirmation bias. 
This is realized through two mutual learning mechanisms elaborated below. 

\subsubsection{Cross-model image perturbation based mutual learning}
\label{sec:CMEMS_cmip}
Given a set of unlabeled images $\mathcal{D}_{U} = {(I_{u}^t)}_{t=1}^{T}$, 
we conduct cross-model mutual learning at the image granularity
by deploying week and strong perturbations, 
defined as $\mathcal{W}$ and $\mathcal{S}$, respectively.
Firstly, we apply the weak perturbation $\mathcal{W}$ to an unlabeled image $I_{u}$ twice 
to generate two weakly augmented images
$\{ I^{\mathcal{W}}_{u,m}: m \in \{1,2\}\}$, which are then separately fed into the two segmentation networks $\mathcal{N}_{m} (m \in \{1,2\})$ to obtain 
probabilistic predictions $P^{\mathcal{W}}_{u,m} \in [0,1]^{K\times H\times W}$ ($m \in \{1,2\}$), 
such that:
\begin{equation}
P^{\mathcal{W}}_{u,m}=\softmax(\mathcal{N}_{m}(I^{\mathcal{W}}_{u,m})),
\label{weakpred}
\end{equation}
where $\softmax$ represents a class-wise softmax function that is used to compute the probability of each pixel belonging to one of the $K$ categories.
Next, we filter out the low-probability predictions and keep only the high-probability ones, 
ensuring that the obtained pseudo-labels $Y^{\mathcal{W}}_{u,m}\in \mathbb{R}^{1\times H\times W}$ consists only of 
confident predictions:
\begin{equation}
Y^{\mathcal{W}}_{u,m}=\argmax(P^{\mathcal{W}}_{u,m})\cdot 
	\mathbb{I}\left[\sum\nolimits_{dim=0}\mathbb{I}\left[P^{\mathcal{W}}_{u,m} \ge \tau\right]\right],
\label{pseudo}
\end{equation}
where $\tau$ represents a predefined threshold used to filter out noisy pseudo-labels;
$\sum_{dim=0}$ indicates summation over the first dimension of the 3D matrix, 
and $\mathbb{I}[\cdot]$ denotes the indicator function.
Then, we apply the strong perturbation $\mathcal{S}$ to each
weak perturbed unlabeled image $I^{\mathcal{W}}_{u,m}$ 
to generate
$I^{\mathcal{S}}_{u,m}$. 
Each $I^{\mathcal{S}}_{u,\bar{m}}$ 
is cross-input into the 
segmentation network  $\mathcal{N}_{m}$ to
produce predictions 
$P^{\mathcal{S}}_{u,m} \in [0,1]^{K\times H\times W}$:
\begin{equation}
	P^{\mathcal{S}}_{u,m}=\softmax(\mathcal{N}_{m}(I^{\mathcal{S}}_{u,\bar{m}})),
\label{strongpred}
\end{equation}
where $\bar{m}\in\{1,2\}$ denotes the complement of $m\in\{1,2\}$ in the set $\{1,2\}$.
Subsequently, 
the cross-model image perturbation consistency loss function is formulated over 
the predictions produced by the two segmentation networks for 
the perturbed unlabeled images as follows:
\begin{equation}
	L_{cmip}= \sum_{m=1}^M L_{seg}(P^{\mathcal{S}}_{u,m},Y^{\mathcal{W}}_{u,\bar{m}})
\label{cmip_loss}
\end{equation}
where $M$ denotes the number of segmentation networks such as $M=2$, 
and $\mathcal{L}_{seg}$ 
denotes the segmentation loss function commonly used in medical image segmentation that
includes both a cross-entropy loss function $L_{ce}$ and a Dice loss function $L_{dice}$: 
\begin{equation}
	\mathcal{L}_{seg}(P,Y)=\frac{1}{2} L_{ce}(P,Y)+\frac{1}{2} L_{dice}(P,Y).
\label{eqseg}
\end{equation}
Here
$P$ represents the predictions and $Y$ indicates the target labels. 
With the proposed consistency loss $L_{cmip}$,
by alternately acting as teachers and students for each other, the two segmentation networks can robustly promote cross-model prediction consistency and mitigate confirmation bias,
producing better pseudo-labels 
and enhancing segmentation performance \citep{arazo2020pseudo}.

\begin{algorithm}[h]
  \caption{Training process of the proposed method}
  \label{alg1}
  \begin{algorithmic}[1]
  \STATE \textbf{Input:} $\mathcal{D}_{E} = (I_{e}, Y_{e})$, $\mathcal{D}_{U} = \{(I_{u}^t)\}_{t=1}^{T}$
  \STATE \textbf{Output:} Trained segmentation networks $\mathcal{N}_{m}(m \in \{1,2\})$
  \STATE Generate the  synthetic dataset $\mathcal{D}_{S} = \{(I_{s}^b, Y_{s}^b)\}_{b=1}^{B}$ by Eq. (\ref{transformation})
  \FOR{$iteration=1,MaxIter$}
   \STATE \quad Sample a batch: $(I_{e}, Y_{e}), (I_{s}, Y_{s}), I_u$
  \STATE \quad $(I^{\mathcal{W}}_{e,1}, Y^{\mathcal{W}}_{e,1}),(I^{\mathcal{W}}_{e,2},Y^{\mathcal{W}}_{e,2}) = \mathcal{W}(I_e,Y_e),\mathcal{W}(I_e,Y_e)$;
  \STATE \quad $(I^{\mathcal{W}}_{s,1}, Y^{\mathcal{W}}_{s,1}),(I^{\mathcal{W}}_{s,2},Y^{\mathcal{W}}_{s,2}) = \mathcal{W}(I_s,Y_s),\mathcal{W}(I_s,Y_s)$;
  \STATE \quad $I^{\mathcal{W}}_{u,1},I^{\mathcal{W}}_{u,2}=\mathcal{W}(I_u),\mathcal{W}(I_u)$;
  \STATE \quad $I^{\mathcal{S}}_{u,1},I^{\mathcal{S}}_{u,2}=\mathcal{S}(I^{\mathcal{W}}_{u,1}),\mathcal{S}(I^{\mathcal{W}}_{u,2})$;
	  \STATE \quad $Y^{\mathcal{W}}_{u,1} \stackrel{\mathcal{N}_{1}}{\longleftarrow} I^{\mathcal{W}}_{u,1}, Y^{\mathcal{W}}_{u,2} \stackrel{\mathcal{N}_{2}}{\longleftarrow} I^{\mathcal{W}}_{u,2}$ by Eq.(\ref{weakpred}), Eq.(\ref{pseudo});
  \STATE \quad $P^{\mathcal{S}}_{u,1} \stackrel{\mathcal{N}_{1}}{\longleftarrow} I^{\mathcal{S}}_{u,2},P^{\mathcal{S}}_{u,2} \stackrel{\mathcal{N}_{2}}{\longleftarrow} I^{\mathcal{S}}_{u,1}$ by Eq.(\ref{strongpred});
  \STATE \quad Compute $L_{cmip}$ by Eq.(\ref{cmip_loss});
  \STATE \quad $P^{\mathcal{FW}}_{u,1} \stackrel{\mathcal{N}_{1}}{\longleftarrow} I^{\mathcal{W}}_{u,2},P^{\mathcal{FW}}_{u,2} \stackrel{\mathcal{N}_{2}}{\longleftarrow} I^{\mathcal{W}}_{u,1}$ by Eq.(\ref{weakfeaturepred});
  \STATE \quad Compute $L_{cmfp}$ by Eq. (\ref{cmfp_loss});
  \STATE \quad $P_{e,1} \stackrel{\mathcal{N}_{1}}{\longleftarrow} I^{\mathcal{W}}_{e,1},P_{e,2} \stackrel{\mathcal{N}_{2}}{\longleftarrow} I^{\mathcal{W}}_{e,2}$;
  \STATE \quad $P_{s,1} \stackrel{\mathcal{N}_{1}}{\longleftarrow} I^{\mathcal{W}}_{s,1},P_{s,2} \stackrel{\mathcal{N}_{2}}{\longleftarrow} I^{\mathcal{W}}_{s,2}$;
  \STATE \quad Compute $L_{e}$ and $L_{s}$ by Eq.(\ref{exemplarloss});
  \STATE \quad Compute $L_{total}$ by Eq.(\ref{totalloss});
  \STATE \quad Update prameters of $\mathcal{N}_{m} (m \in \{1,2\})$;
  \ENDFOR
  \end{algorithmic}
\end{algorithm}
\begin{table*}\small
\caption{\label{tab:Synapse_DSC} 
Quantitative comparison results on the Synapse dataset. 
We report the class average DSC and HD95 results and the DSC results for all individual classes.
}
\setlength{\tabcolsep}{1.05mm}
\begin{center}
\begin{tabular}{c|c|ccccccccc}
\hline
\textbf{Method}  & HD95.Avg$\downarrow$&DSC.Avg$\uparrow$ & Aor & Gal&Kid(L)&Kid(R)&Liv&Pan&Spl&Sto\\
\hline
	UNet \citep{ronneberger2015u}&132.42&0.160&0.026&0.167&0.177&0.154&0.649&0.015&0.059&0.033\\
	MTUNet \citep{wang2022mixed}&154.60&0.112&0.066&0.108&0.155&0.053&0.352&0.008&0.046&0.102 \\
	MLDS \citep{reiss2021every}&159.26&0.221&0.057&0.147&0.306&0.183&0.638&0.038&0.306&0.090\\ FixMatch \citep{sohn2020fixmatch}&154.49&0.235&0.009&0.174&0.349&0.126&0.697&0.037&0.403&0.084\\    CPS \citep{chen2021semi}&140.27&0.212&0.014&0.267&0.202&0.153&0.693&0.093&0.225&0.048\\    ELSNet \citep{en2022exemplar}&109.70&0.315&0.319&\textbf{0.372}&0.381&0.219&0.784&0.067&0.276&0.104 \\ CMEMS&\textbf{55.02}&\textbf{0.597}&\textbf{0.840}&0.230&\textbf{0.802}&\textbf{0.757}&\textbf{0.833}&\textbf{0.153}&\textbf{0.873}&\textbf{0.291}\\
    \hline
    \hline
	FullySup &39.70&0.769&0.891&0.697&0.778&0.686&0.934&0.540&0.867&0.756\\
\hline
\end{tabular}
\vskip -.2in
\end{center}
\end{table*}

\subsubsection{Cross-model multi-level feature perturbation based mutual learning}
\label{sec:CMEMS_cmfp}
Although cross-model image perturbation learning can enhance a network's prediction robustness, it can only exploit perturbations that originate in the input space and may obstruct the exploration of consistency in 
various level
features across models.
Therefore, we propose a straightforward yet effective cross-model mutual learning 
through multi-level feature perturbations 
to further enhance consistency in the feature granularity of our segmentation models.
Specifically, we cross-input the weak perturbed unlabeled image 
$I^{\mathcal{W}}_{u,\bar{m}}$ into the encoder $f_m$ 
to generate its multi-level features 
$X^{\mathcal{W}}_{u,\bar{m}}= \{x^{\mathcal{W},l}_{u,\bar{m}}\}^{L=5}_{l=1}$,
where $L=5$ denotes the total number of layers. 
We then apply 
a feature perturbation operation $\mathcal{F}$ to the multi-level features, 
and fed the perturbed features 
into the corresponding layers of 
the decoder $g_{m}$ to obtain the predictions $P^{\mathcal{FW}}_{u,m}$ as follows:
\begin{equation}
	P^{\mathcal{FW}}_{u,m}=\softmax(g_m(\mathcal{F}(X^{\mathcal{W}}_{u,\bar{m}})).
\label{weakfeaturepred}
\end{equation}
The perturbation operation 
$\mathcal{F}$ is defined as 
randomly dropping out channels with a probability of 0.5.
Subsequently, 
we formulate the cross-model multi-level feature perturbation consistency loss over unlabeled images as follows:
\begin{equation}
	L_{cmfp}= \sum_{m=1}^M L_{seg}(P^{\mathcal{FW}}_{u,m},Y^{\mathcal{W}}_{u,\bar{m}}).
\label{cmfp_loss}
\end{equation}
The proposed $L_{cmfp}$ maintains cross-model 
segmentation consistency with broadened perturbation space, 
strengthening the robustness of our framework.
Moreover, the multi-level feature perturbation
based learning 
is performed on features with different resolutions.
It allows the framework to capture 
useful semantic 
information at various depths 
and complement our cross-model image perturbation based
learning with various levels of detail,
enhancing the segmentation models.

\subsection{Collaborative Optimization Procedure}
\label{sec:training}
The proposed CMEMS is trained end-to-end.
In each iteration, we sample a batch of images,
which includes
the annotated exemplar image $(I_{e}, Y_{e})$,
an annotated synthetic image $(I_{s}^b, Y_{s}^b)$,
and an unlabeled image $I_{u}^t$,
to update the segmentation networks by minimizing a joint loss function. 
Specifically, the overall loss function for training the segmentation networks $\mathcal{N}_{m} (m \in \{1,2\})$ contains four terms:
\begin{equation}
L_{total} = 
	L_{e}+L_{s}+\lambda_{cmip}L_{cmip}+\lambda_{cmfp}L_{cmfp},
\label{totalloss}
\end{equation}
where 
$\lambda_{cmip}$ and $\lambda_{cmfp}$ are trade-off hyperparameters.
$L_{e}$ and $L_{s}$ represent an exemplar segmentation loss and a synthetic segmentation loss calculated from the exemplar and synthetic images, respectively:
\begin{equation}
L_{e}=\sum_{m=1}^{M}\mathcal{L}_{seg}(P_{e,m},Y_{e}),\;\;
L_{s}=\sum_{m=1}^{M}\mathcal{L}_{seg}(P_{s,m},Y_{s}),
\label{exemplarloss}
\end{equation}
where $P_{e,m}$ and $P_{s,m}$ denote the predictions of $\mathcal{N}_{m}$
on the exemplar and synthetic images, respectively.
$L_{cmip}$ indicates the cross-model image perturbation consistency loss defined in Eq.(\ref{cmip_loss}), and $L_{cmfp}$ is the cross-model multi-level feature perturbation consistency loss defined in Eq.(\ref{cmfp_loss}).
The training process of the proposed framework is described in Algorithm \ref{alg1}.

\section{EXPERIMENTS}
\label{sec:experimental}
\subsection{Experimental Settings}
\paragraph{Datasets and evaluation metrics}
The proposed CMEMS framework is evaluated on the Synapse multi-organ CT dataset and the Automated Cardiac Diagnosis Challenge (ACDC) dataset. 
The Synapse dataset comprises 30 abdominal CT cases containing 2,211 images with eight abdominal organs. 
We followed the experimental setup of \citet{wang2022mixed} and \citet{chen2021transunet} 
and used 18 cases for training and 12 cases for testing. 
The ACDC dataset contains 100 cardiac MRI cases with 1,300 images, and each image is labeled 
with three organ categories. 
We used 70 cases for training, 20 for validation, and 10 for testing. 
We used the DSC and the HD95 as evaluation metrics \citep{wang2022mixed,chen2021transunet} to assess the performance of the proposed CMEMS framework.

\begin{table*}[t!]\small
\caption{\label{tab:ACDC}
Quantitative comparison results on the ACDC dataset.
We report the class average results and the results for individual classes in terms of DSC and HD95.
}
\begin{center}
\setlength{\tabcolsep}{1.2mm}
\begin{tabular}{c|cccc|cccc}
\hline
Method&DSC.Avg$\uparrow$&RV&Myo&LV&HD95.Avg$\downarrow$&RV&Myo&LV\\
\hline
    UNet \citep{ronneberger2015u}& 0.142 &0.140 &0.112&0.174&43.30&63.76&35.60&30.80\\
    MT-UNet \citep{wang2022mixed}& 0.142&0.119&0.126&0.182&74.20&83.91&61.48&77.22\\
    MLDS \citep{reiss2021every} &0.189&0.144&0.165&0.258&50.03&72.13&30.20&47.77\\   FixMatch \citep{sohn2020fixmatch}&0.529&0.291&0.606&0.691&43.18&85.80&11.02&32.73\\
    CPS \citep{chen2021semi}&0.194&0.130&0.180&0.271&66.08&85.12&62.89&50.23\\ ELSNet \citep{en2022exemplar}&0.410&0.293&0.374&0.563&26.64&47.63&16.58&15.73\\    CMEMS&\textbf{0.817}&\textbf{0.759}&\textbf{0.793}&\textbf{0.900}&\textbf{7.35}&\textbf{12.91}&\textbf{3.48}&\textbf{5.67}\\
    \hline\hline
	FullySup &0.898&0.882&0.883&0.930&7.00&6.90&5.90&8.10\\
\hline
\end{tabular}
\end{center}
\vskip -.2in
\end{table*}

\paragraph{Implementation details}
We adopt the UNet \citep{ronneberger2015u} as our base network 
due to its
effectiveness and efficiency in medical image segmentation.
We randomly initialize the weights of the two
segmentation networks, 
$\mathcal{N}_1$ and $\mathcal{N}_2$,
and resize the input images to 224$\times$224. 
We define the weak perturbation operation $\mathcal{W}$ as random rotation and flipping, 
and define the strong perturbation operation $\mathcal{S}$ as color jittering with an intensity factor $\alpha$. 
We set the batch size to 12, set $\tau$ to 0.8, and set $\alpha$ 
to 1.0 and 0.2 on ACDC and Synapse, respectively. 
For the ACDC dataset, we set $\lambda_{cmip}$ and $\lambda_{cmfp}$ to 1 and 0.09, respectively, while for the Synapse dataset, we set them to 0.1 and 0.09, respectively. 
To optimize the proposed framework, we use the Adam optimizer with a weight decay of 0.0001 and a learning rate of 3e-4. 
We use the same settings as ELSNet \citep{en2022exemplar} for the synthetic dataset, \ie, ten synthetic images for the same background on the Synapse dataset and fifteen on the ACDC dataset.
For evaluation, we followed the experimental setup described in the literature \citep{chen2021transunet}, where all 3D volumes are split into individual images for inference. 
Our experiments are 
conducted using NVIDIA GTX 2080Ti GPU.

\begin{table*}\small
\renewcommand\arraystretch{1.1}
\caption{\label{tab:ab_Synapse_DSC_avg} 
Ablation study of the proposed components on the ACDC and Synapse datasets.
We report the class average DSC and HD95 results.
SD: using synthetic dataset.
CM: cross-model mutual learning.
IP: using image perturbations.
FP: using multi-level feature perturbations.
Data: datasets used in the training process.
}
\begin{center}
\begin{tabular}{c|cccc|cc|cc}
\hline
\multirow{2}*{Data}&&&&&\multicolumn{2}{c|}{Synapse}&\multicolumn{2}{c}{ACDC}\\
&SD&CM&IP&FP &DSC.Avg$\uparrow$  & HD95.Avg$\downarrow$&DSC.Avg$\uparrow$  & HD95.Avg$\downarrow$ \\
\hline
	$\mathcal{D}_{E}$&-&-&-&-&0.160&132.42&0.142&43.30\\
 $\mathcal{D}_{E}$+$\mathcal{D}_{S}$&$\surd$&-&-&-&0.256&122.49&0.359&15.16\\
    $\mathcal{D}_{E}$+$\mathcal{D}_{U}$&-&$\surd$&-&-&0.212&140.27&0.194&66.08\\ 
    $\mathcal{D}_{E}$+$\mathcal{D}_{U}$&-&-&$\surd$&-&0.235&154.49&0.529&43.18\\
    $\mathcal{D}_{E}$+$\mathcal{D}_{S}$+$\mathcal{D}_{U}$&$\surd$&$\surd$&-&-&0.378&113.07&0.516&10.43\\
    $\mathcal{D}_{E}$+$\mathcal{D}_{S}$+$\mathcal{D}_{U}$&$\surd$&-&$\surd$&-&0.429&106.76&0.670&20.67\\
    $\mathcal{D}_{E}$+$\mathcal{D}_{S}$+$\mathcal{D}_{U}$&$\surd$&$\surd$&$\surd$&-&0.523&101.93&0.807&8.07\\
    $\mathcal{D}_{E}$+$\mathcal{D}_{S}$+$\mathcal{D}_{U}$&$\surd$&$\surd$&$\surd$&$\surd$&\textbf{0.597}&\textbf{55.02}&0.817&7.35\\
\hline
\end{tabular}
\vskip -.2in
\end{center}
\end{table*}

\begin{table}[t]
\vskip -.2in
\renewcommand\arraystretch{1.3}
\caption{\label{tab:ablation_weakperturbation}
Ablation study on using different or same weak perturbations for the two segmentation networks on the Synapse and ACDC datasets.
}
\begin{center}\small
\begin{tabular}{c|cc|cc}
\hline
\multirow{2}*{Method}&\multicolumn{2}{c|}{Synapse}&\multicolumn{2}{c}{ACDC}\\
&DSC$\uparrow$&HD95$\downarrow$&DSC$\uparrow$&HD95$\downarrow$\\
\hline
Same $I^{\mathcal{W}}_{e,m}$ &0.550&74.25&0.805&6.84\\
\hline
Same $I^{\mathcal{W}}_{s,m}$ &0.542&80.73&0.800&6.93\\
\hline
Same $I^{\mathcal{W}}_{u,m}$ &0.583&71.36&0.812&6.94\\
\hline
Different &0.597&55.02&0.817&6.35\\
\hline
\end{tabular}
\end{center}
\vskip -.2in
\end{table}

\subsection{Quantitative Evaluation Results}
The proposed CMEMS is compared with six existing state-of-the-art methods on the Synapse and ACDC datasets.
To ensure a fair comparison, we utilize the same  exemplar, synthetic datasets, unlabeled datasets, backbone architecture and perturbation operations for all the comparison methods. 

\paragraph{Comparison results on Synapse}
The test results of the proposed CMEMS and the other comparison methods on the Synapse dataset are reported in Table \ref{tab:Synapse_DSC}.
The results show that CMEMS achieves superior performance, outperforming all the other compared methods in terms of both DSC and HD95 metrics.
MLDS \citep{reiss2021every}, FixMatch \citep{sohn2020fixmatch}, and CPS \citep{chen2021semi} are semi-supervised methods 
without effective integration of complementary information among models and perturbation information from multiple levels. 
The proposed CMEMS surpasses these methods 
in terms of class average DSC
by 0.376, 0.362, and 0.385, respectively.
Besides, the proposed CMEMS outperforms the second-best method, ELSNet \citep{en2022exemplar}, 
by 0.282 in terms of class average DSC, 
while reducing the class average HD95 value from 109.7 to 55.02.
Moreover, Gal, Pan and Sto exhibit small organ areas and distinctive shape variations, 
resulting in general lower performance. 
Nevertheless, our method still works better than most existing methods in these categories.
These experimental results illustrate the effectiveness of the proposed CMEMS.

\paragraph{Comparison results on ACDC}
The test results of the proposed CMEMS and the other comparison methods on the ACDC dataset are reported in Table \ref{tab:ACDC}.
The results indicate that the proposed CMEMS produces substantial improvements 
over other methods in terms of both DSC and HD95 metrics, 
achieving the best class average DSC and HD95 scores of 0.817 and 7.35, respectively.
FixMatch \citep{sohn2020fixmatch} is the second-best-performing method, yet it performs poorly in terms of HD95.
Although ELSNet \citep{en2022exemplar} performs the second best in terms of HD95, it does not perform well in terms of DSC. 
Surprisingly, the performance of our proposed CMEMS is close to that of the fully supervised method. 
These findings again demonstrate the effectiveness of CMEMS for exemplar-based medical image segmentation.

\subsection{Ablation Study}
\paragraph{Impact of different components}
To investigate the contributions of each component to the 
performance of CMEMS,
we conducted a set of experiments on Synapse and ACDC. 
The results are reported in Table \ref{tab:ab_Synapse_DSC_avg}.
On Synapse, the experimental results (from the second to the fourth row) show that using the synthetic dataset (SD), 
conducting cross-model mutual learning without perturbations (CM), 
and using image perturbations (IP) 
can improve the DSC result 
from the base 0.16 to 0.256, 0.212, and 0.235, respectively.
Employing the synthetic dataset (SD) together with CM or IP can further improve the experimental results,
while substantial improvements can be obtained by 
considering both synthetic and unlabeled data and 
combining both CM and IP, as shown in the second last row.
By further taking the multi-level feature perturbations (FP) into consideration,
the full model with all of the components yields the best results in terms of both DSC and HD95.  
Similar observations can be made on the ACDC dataset. 
In summary, the experiment results illustrated the effectiveness of each component.

\paragraph{Impact of using different weak perturbations}
We summarize the impact of using different
weak image perturbations 
as inputs for the two segmentation networks 
in Table \ref{tab:ablation_weakperturbation}. 
The first three rows indicate that we applied a single weak perturbation to the exemplar image ($I_e$), synthetic image ($I_s$), and unlabeled image ($I_u$), respectively, before feeding them into the two segmentation networks, 
\ie, $I^{\mathcal{W}}_{d,1}=I^{\mathcal{W}}_{d,2}$ with $d \in \{e,s,u\}$. 
The results show that the best segmentation performances 
on both datasets are obtained when using 
two different
weak perturbations on each input image to produce inputs for the two networks, as shown in the last row.
In contrast, segmentation performance is degraded
when the same weak perturbations are deployed on any type of images. 
We attribute this to the fact that applying 
different weak perturbations to each of the input images, 
together with our proposed cross-model perturbation based mutual learning, 
can significantly increase the diversity
of training data, leading to more robust segmentation networks.

\begin{table}[t]
\vskip -.2in
\renewcommand\arraystretch{1.3}
\caption{\label{tab:ablation_MLFP}
Ablation study of cross-model versus single-model multi-level feature perturbations on Synapse and ACDC in terms of DSC and HD95.
}
\begin{center}\small
\begin{tabular}{c|cc|cc}
\hline
\multirow{2}*{Method}&\multicolumn{2}{c|}{Synapse}&\multicolumn{2}{c}{ACDC}\\
&DSC$\uparrow$&HD95$\downarrow$&DSC$\uparrow$&HD95$\downarrow$\\
\hline
Individual-model &0.587&66.51&0.812&6.53\\
\hline
Cross-model &0.597&55.02&0.817&6.35\\
\hline
\end{tabular}
\end{center}
\vskip -.2in
\end{table}

\begin{figure}
\centering
  \includegraphics[width=0.9\linewidth,height=30mm]{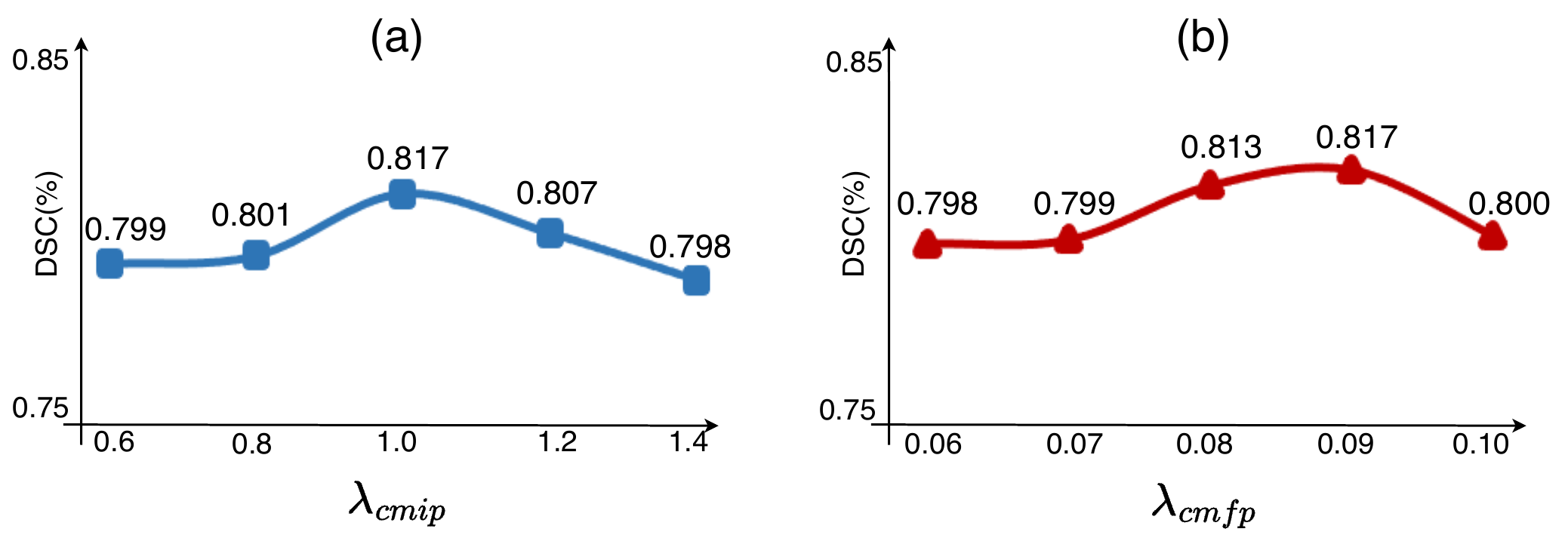}\\
\caption{
	Impact of the weight (a) $\lambda_{cmip}$ and (b) $\lambda_{cmfp}$ on 
	the performance on the ACDC dataset.
}
  \label{fig:weight_ab}
\end{figure}

\begin{figure}
\centering
  \includegraphics[width=0.9\linewidth,height=30mm]{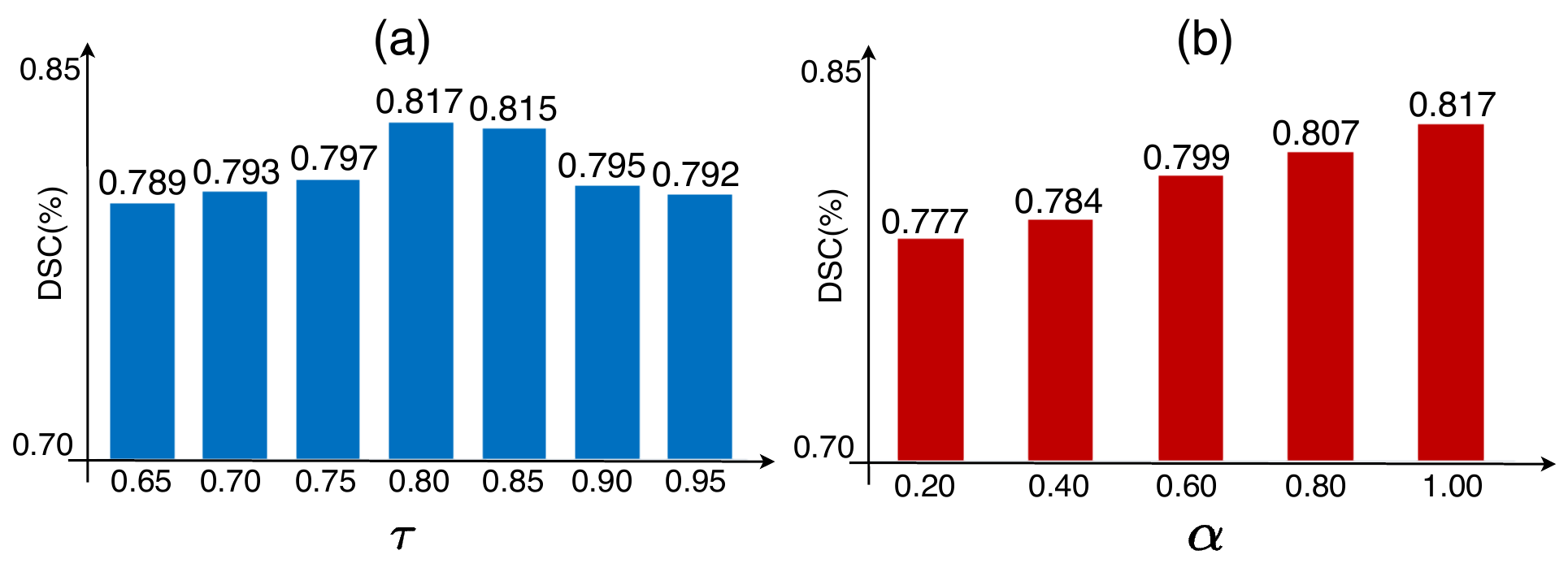}\\
\caption{
	Impact of (a) the confidence threshold $\tau$,  and (b) the intensity factor $\alpha$ 
	on the performance on the ACDC dataset.
}
  \label{fig:threshold_ab}
\end{figure}

\paragraph{Impact of cross-model versus single-model multi-level feature perturbation}
We summarize the effect of cross-model versus single-model multi-level feature perturbations 
on the Synapse and ACDC datasets, as shown in Table \ref{tab:ablation_MLFP}.
The ``Individual-model'' variant indicates that we input $I^{W}_{u,1}$ into $\mathcal{N}_{1}$ to obtain $P^{\mathcal{FW}}_{u,2}$, and input $I^{W}_{u,2}$ into $\mathcal{N}_{2}$ to obtain $P^{\mathcal{FW}}_{u,1}$.
In this case, the two models can independently 
use their own pseudo-labels for the feature perturbed predictions through the loss term $L_{cmfp}$.  
The results indicate that using the cross-model feature perturbation based predictions
is preferable than using the individual single-model based predictions, 
improving the DSC results on the Synapse and the ACDC datasets by 1\% and 0.5\%, respectively.
These experimental results validate that the cross-model feature perturbation
based learning is effective.

\begin{figure*}
\centering
  \includegraphics[width=0.85\linewidth,height=85mm]{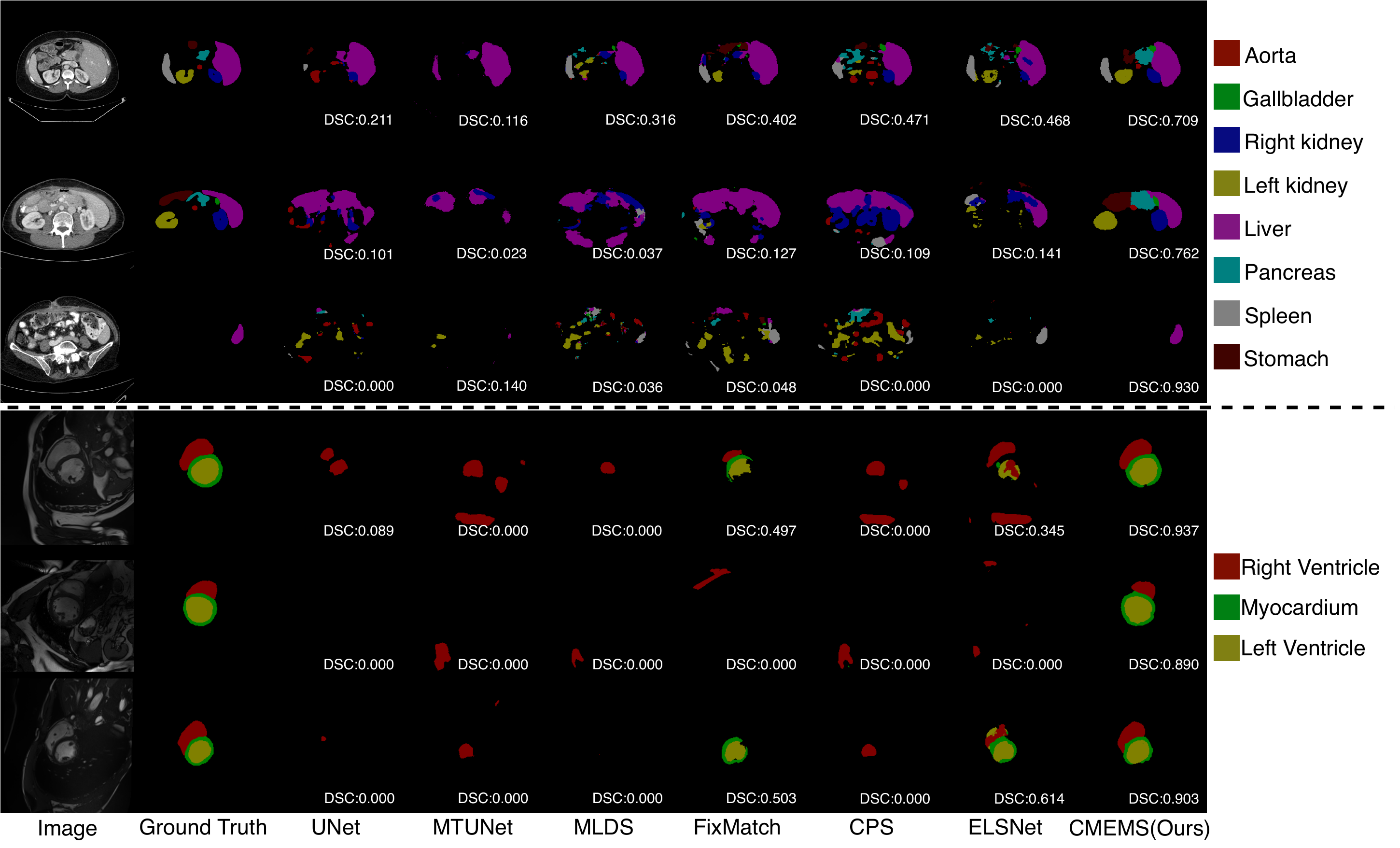}\\
\caption{
Visual examples of the segmentation results obtained by the proposed CMEMS framework and other state-of-the-art methods on the Synapse and ACDC datasets. 
The first two columns display the input images and the corresponding ground truth labels. 
The last column shows visualization of the segmentation results generated by CMEMS. 
The remaining columns show the results obtained by other methods.
}
  \label{fig:visualization}
\end{figure*}

\subsection{Parameter Analysis}
\paragraph{Impact of the weights of mutual learning loss functions}
\label{sec:weight_param}
We summarize the impact of the values of $\lambda_{cmip}$ and $\lambda_{cmfp}$ on the model performance 
on the ACDC dataset in Figure \ref{fig:weight_ab}.
The results in Figure \ref{fig:weight_ab} (a) show that the best performance is obtained when $\lambda_{cmip}$ is set to 1, achieving a DSC value of 0.817,
while decreasing or increasing $\lambda_{cmip}$ leads to performance degradation.
This suggests that it is important to exploit the unlabeled images through cross-model image perturbation based mutual learning, 
but overly emphasizing the unlabeled loss $L_{cmip}$ is not desirable. 
In Figure \ref{fig:weight_ab} (b), it is evident that 
the performance improves gradually
when $\lambda_{cmfp}$ increases from 0.06 to 0.09,
and then degrades when the $\lambda_{cmfp}$ value further increases to 0.1. 
This indicates that a relatively small $\lambda_{cmfp}$ value leads to desirable results, 
suggesting that $L_{cmfp}$ should only be used as a slightly weighted auxiliary loss. 

\paragraph{Impact of confidence threshold $\tau$}
\label{sec:conﬁdence_param}
We conducted experiments on ACDC to 
investigate the impact of the confidence threshold $\tau$ on the model performance 
and report the results in Figure \ref{fig:threshold_ab} (a).
The parameter $\tau$ plays a crucial role in balancing the quality and quantity of 
the generated pseudo-labels. An optimal value of 0.8 for $\tau$ yields the 
best segmentation result of 0.817 in terms of DSC. 
Decreasing or increasing $\tau$ to 0.65 or 0.95 results in reduced 
experimental performance. 
This underscores the significance 
of both the quality and quantity of the pseudo-labels 
for learning effective segmentation models. 

\paragraph{Impact of intensity factor $\alpha$}
\label{sec:intensity_param}
We summarize the impact of the intensity factor $\alpha$ for strong perturbations (\ie, colour jittering) 
over model performance
on the ACDC dataset in Figure \ref{fig:threshold_ab} (b). 
The $\alpha$ value controls
the range of varying intensity of brightness, contrast and saturation.
The experimental results indicate that a larger intensity factor leads to better results, 
with the best results obtained when $\alpha$=1.0, reaching a DSC value of 0.817. 
This suggests that 
a wider range of colour jittering operations could 
help with cross-model mutual learning on unlabeled data. 

\subsection{Qualitative Evaluation Results}
To demonstrate the effectiveness of the proposed method, we present the visualization comparisons with existing state-of-the-art methods in Figure \ref{fig:visualization}.
The results demonstrate that the proposed CMEMS outperforms all the other methods in terms of visual segmentation on both datasets.
The background clutter and the low brightness phenomenon in medical images can greatly affect the segmentation results of 
the existing methods, making them prone to misclassifying background regions as foreground organs.
Surprisingly, the proposed CMEMS can obtain segmentation results nearly as good as the ground-truth, 
benefiting from its ability to mitigate confirmation bias and learn complementary information.

\section{CONCLUSION}
In this paper, we proposed a novel framework, CMEMS, to achieve exemplar-based medical image segmentation by utilizing 
two mutual learning models 
to excavate implicit information from unlabeled data at multiple granularities. 
The CMEMS enables the cross-model image perturbation based mutual learning by using pseudo-labels generated 
by one model from weakly perturbed images 
to supervise predictions of the other model over strongly perturbed images. 
Moreover, the cross-model multi-level feature perturbation 
based mutual learning is designed to broaden the perturbation space and further enhance the robustness of the proposed framework.
The experimental results demonstrate that the proposed CMEMS substantially outperforms the state-of-the-art methods.

{\small
\bibliographystyle{apalike}
\bibliography{aistats24}
}

\newpage
\onecolumn
\appendix

\section{Additional Ablation Study Results}
\label{sec:Additional Ablation Studies}
We present the detailed ablation study results 
regarding the model components
on the Synapse and ACDC datasets 
in Table \ref{tab:ab_Synapse_DSC} and Table \ref{tab:ab_ACDC_DSC}, respectively.
Table \ref{tab:ab_Synapse_DSC} demonstrates substantial enhancements in the DSC values 
for each category on the Synapse dataset with the incorporation of the various components. 
Notably, even for Aor, the organ with the smallest area in this dataset, the DSC value exhibits remarkable improvements, increasing from 0.026 to a final value of 0.840. 
A similar noteworthy enhancement is observed for the Spl category as well. 
These results highlight the effectiveness of CMEMS in significantly improving 
the segmentation outcomes,
particularly for challenging categories.
Furthermore, Table \ref{tab:ab_ACDC_DSC} reveals significant improvements in terms of both DSC and HD95 metrics for each category 
of the ACDC dataset.

\begin{table*}[h]\small
\renewcommand\arraystretch{1.1}
\caption{\label{tab:ab_Synapse_DSC} 
Ablation study of the proposed components on the Synapse dataset.
We report the class average DSC and HD95 results and the DSC results for all individual classes.
SD: using synthetic dataset.
CM: cross-model mutual learning.
IP: using image perturbations.
FP: using multi-level feature perturbations.
Data: datasets used in the training process.
}
\setlength{\tabcolsep}{1.2mm}
\begin{center}
\begin{tabular}{c|cccc|c|ccccccccc}
\hline
Data&SD&CM&IP&FP  & HD95.Avg$\downarrow$&DSC.Avg$\uparrow$ & Aor & Gal&Kid(L)&Kid(R)&Liv&Pan&Spl&Sto\\
\hline
	$\mathcal{D}_{E}$&-&-&-&-&132.42&0.160&0.026&0.167&0.177&0.154&0.649&0.015&0.059&0.033\\
 $\mathcal{D}_{E}$+$\mathcal{D}_{S}$&$\surd$&-&-&-&122.49&0.256&0.112&0.273&0.321&0.129&0.792&0.008&0.360&0.051\\
    $\mathcal{D}_{E}$+$\mathcal{D}_{U}$&-&$\surd$&-&-&140.27&0.212&0.014&0.267&0.202&0.153&0.693&0.093&0.225&0.048\\ 
    $\mathcal{D}_{E}$+$\mathcal{D}_{U}$&-&-&$\surd$&-&154.49&0.235&0.009&0.174&0.349&0.126&0.697&0.037&0.403&0.084\\
    $\mathcal{D}_{E}$+$\mathcal{D}_{S}$+$\mathcal{D}_{U}$&$\surd$&$\surd$&-&-&113.07&0.378&0.658&0.328&0.510&0.079&0.689&0.141&0.526&0.094\\
    $\mathcal{D}_{E}$+$\mathcal{D}_{S}$+$\mathcal{D}_{U}$&$\surd$&-&$\surd$&-&106.76&0.429&0.700&0.256&0.499&0.185&0.826&0.143&0.533&0.289\\
    $\mathcal{D}_{E}$+$\mathcal{D}_{S}$+$\mathcal{D}_{U}$&$\surd$&$\surd$&$\surd$&-&101.93&0.523&0.783&\textbf{0.414}&0.585&0.468&0.828&0.027&0.695&\textbf{0.380}\\
    $\mathcal{D}_{E}$+$\mathcal{D}_{S}$+$\mathcal{D}_{U}$&$\surd$&$\surd$&$\surd$&$\surd$&\textbf{55.02}&\textbf{0.597}&\textbf{0.840}&0.230&\textbf{0.802}&\textbf{0.757}&\textbf{0.833}&\textbf{0.153}&\textbf{0.873}&0.291\\
\hline
\end{tabular}
\end{center}
\end{table*}
\begin{table*}[h]\small
\renewcommand\arraystretch{1.1}
\caption{\label{tab:ab_ACDC_DSC} 
Ablation study of the proposed components on the ACDC dataset.
We report the class average DSC and HD95 results and the DSC and HD95 results for all individual classes.
CM: cross-model mutual learning.
IP: using image perturbations.
FP: using multi-level feature perturbations.
Data: datasets used in the training process.
}
\begin{center}
\begin{tabular}{c|cccc|cccc|cccc}
\hline
Data&SD&CM&IP&FP  & DSC.Avg$\uparrow$&RV&Myo&LV&HD95.Avg$\downarrow$&RV&Myo&LV\\
\hline
	$\mathcal{D}_{E}$&-&-&-&-&0.142 &0.140 &0.112&0.174&43.30&63.76&35.60&30.80\\
 $\mathcal{D}_{E}$+$\mathcal{D}_{S}$&$\surd$&-&-&-&0.359&0.193&0.347&0.535&15.16&21.19&10.98&13.32\\
 $\mathcal{D}_{E}$+$\mathcal{D}_{U}$&-&$\surd$&-&-&0.194&0.130&0.181&0.272&66.08&85.13&62.90&50.23\\ 
    $\mathcal{D}_{E}$+$\mathcal{D}_{U}$&-&-&$\surd$&-&0.529&0.291&0.606&0.691&43.18&85.80&11.02&32.73\\   
    $\mathcal{D}_{E}$+$\mathcal{D}_{S}$+$\mathcal{D}_{U}$&$\surd$&$\surd$&-&-&0.516&0.284&0.572&0.693&10.43&26.98&2.50&1.83\\
    $\mathcal{D}_{E}$+$\mathcal{D}_{S}$+$\mathcal{D}_{U}$&$\surd$&-&$\surd$&-&0.670&0.618&0.611&0.781&20.67&27.71&16.66&17.64\\
    $\mathcal{D}_{E}$+$\mathcal{D}_{S}$+$\mathcal{D}_{U}$&$\surd$&$\surd$&$\surd$&-&0.807&0.737&0.785&0.900&8.07&13.05&5.10&6.05\\
    $\mathcal{D}_{E}$+$\mathcal{D}_{S}$+$\mathcal{D}_{U}$&$\surd$&$\surd$&$\surd$&$\surd$&\textbf{0.817}&\textbf{0.759}&\textbf{0.793}&\textbf{0.900}&\textbf{7.35}&\textbf{12.91}&\textbf{3.48}&\textbf{5.67}\\
\hline
\end{tabular}
\end{center}
\end{table*}

\begin{table}[t]
\renewcommand\arraystretch{1.3}
\caption{\label{tab:ablation_diffexemp}
Results with different exemplars on ACDC in terms of DSC.Avg.
}
\begin{center}\small
\begin{tabular}{c|c|c|c}
\hline
& Exemplar1 & Exemplar2 & Exemplar3\\
\hline
UNet \citep{ronneberger2015u}  & 0.160 & 0.114 & 0.145 \\
\hline
ELSNet \citep{en2022exemplar}  & 0.410 & 0.399 & 0.409 \\
\hline
CMEMS(Ours)  & 0.817 & 0.809 & 0.812  \\
\hline
\end{tabular}
\end{center}
\end{table}

\begin{table}[h]
\renewcommand\arraystretch{1.3}
\caption{\label{tab:ablation_sp}
Results with different types of strong perturbations on ACDC in terms of DSC.Avg.
CJ: color jittering.
GB: Gaussian blur.
AS: sharpness adjustment.
}
\begin{center}\small
\begin{tabular}{c|c|c|c}
\hline
 & CJ & CJ+GB & CJ+GB+AS\\
\hline
CMEMS(Ours)  & 0.817 & 0.818 & 0.818  \\
\hline
\end{tabular}
\end{center}
\end{table}

\section{Additional Parameter Analysis Results}
\label{sec:Additional Parameter Analysis}

\subsection{Impact of different exemplars}
\label{sec:different_exemplars}
We summarize the 
results with 
different exemplars on the ACDC dataset in terms of DSC values in Table \ref{tab:ablation_diffexemp}.
We randomly selected three exemplar samples
and compared 
our proposed CMEMS with UNet and ELSNet. 
The results indicate that the performance of our proposed CMEMS is more stable 
than that of the other methods when 
using different exemplars, 
and CMEMS substantially outperforms the other two compared methods regardless of the exemplar used.
In particular, ELSNet uses the same data as our CMEMS,
while CMEMS outperforms ELSNet by more than 0.4 in terms of class average DSC. 

\subsection{Impact of different types of strong perturbations}
\label{sec:different_perturbation}
We summarize the results 
with different types of strong perturbations on the ACDC dataset in terms of DSC values in Table \ref{tab:ablation_sp}.
In addition to colour jittering,
we tested Gaussian blur and sharpness adjustment.
The experimental results show that 
the performance of our proposed method 
does not change much 
by adding more strong perturbation methods. 
It also validates that using colour jittering is sufficient.

\begin{figure}
\centering
  \includegraphics[width=0.4\linewidth,height=50mm]{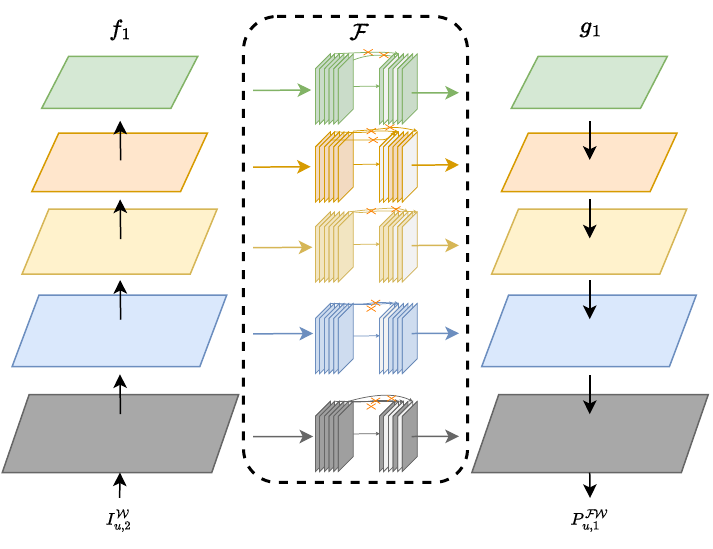}\\
\caption{
The architecture of multi-level feature perturbation.
}
  \label{fig:mlfp}
\end{figure}

\begin{table}[h]
\small
\renewcommand\arraystretch{1.3}
\caption{\label{tab:SSL} 
Additional quantitative comparison results on the ACDC dataset. 
}
\setlength{\tabcolsep}{1.1mm}
\begin{center}
\begin{tabular}{c|ccc}
\hline
Method&\citep{chaitanya2020contrastive}&\citep{chaitanya2020contrastive}+Mixup&CMEMS(Ours) \\
\hline
Num of labels& $\sim$ 100& $\sim$ 100& 1\\
\hline
DSC.Avg&0.725&0.757&\textbf{0.817}\\
\hline
\end{tabular}
\end{center}
\end{table}

\section{Additional Quantitative Evaluation Results}
\label{sec:Additional Quantitative Evaluation Results}
To further illustrate the efficacy of our proposed CMEMS, we conducted a comparison with a self-supervised learning approach for semi-supervised medical image segmentation \citep{chaitanya2020contrastive}. 
The results are presented in Table \ref{tab:SSL},
which show that 
our CMEMS method outperforms the compared approach when evaluated on the ACDC dataset. 
Our proposed CMEMS achieves superior results even when trained with only a single 
labeled image (i.e., the exemplar), surpassing the performance of the compared method trained with approximately 100 
labeled images. 
Moreover, our method surpasses the performance of the compared method augmented with the Mixup technique, further underscoring the robustness and effectiveness of the proposed CMEMS in the context of medical image segmentation.

\section{Details of the Model Architecture}
\label{sec:details_architecture}
Our base network is the UNet, which contains an encoder and a decoder.
The encoder contains 1*ConvBlock and 4*DownBlock.
ConvBlock: Conv3*3 $\to$ BatchNorm $\to$ LeakyReLU $\to$ Dropout $\to$ Conv3*3 $\to$ BatchNorm $\to$ LeakyReLU.
DownBlock: MaxPool2*2 $\to$ ConvBlock.
The decoder contains 4*UpBlock and 1*Final Layer.
UpBlock:  Conv1*1 $\to$ Upsample $\to$ Concat $\to$ ConvBlock.
Final Layer: Conv3*3.
The encoder feature channels and dropout rates are \{16, 32, 64, 128, 256\} and \{0.05, 0.1, 0.2, 0.3, 0.5\}, while the decoder has \{256, 128, 64, 32, 16\} channels and a dropout rate of 0.

In addition, we present the architecture of the multi-level feature perturbation operation 
on UNet in Figure \ref{fig:mlfp}, where  
the decoder takes the multi-level feature outputs from the encoder as inputs at the corresponding levels.

\end{document}